\documentclass[twoside]{article}

 \usepackage[accepted]{aistats2020}

\usepackage[round]{natbib}

\usepackage[utf8]{inputenc}
\usepackage{url}
\RequirePackage[colorlinks=true,citecolor=blue,linkcolor=blue,urlcolor=blue]{hyperref}

\usepackage{amsfonts}
\usepackage{amsmath, amssymb, amsthm}
\usepackage{graphicx, xcolor}
\usepackage{booktabs}
\usepackage[nameinlink]{cleveref}

\newtheorem{theorem}{Theorem}
\newtheorem{lemma}[theorem]{Lemma}
\newtheorem{proposition}[theorem]{Proposition}

\crefname{example}{Example}{examples}
\crefname{theorem}{Theorem}{Theorems}
\crefname{lemma}{Lemma}{Lemmas}
\crefname{proposition}{Proposition}{Propositions}
\crefname{remark}{Remark}{Remarks}
\crefname{corollary}{Corollary}{Corollaries}
\crefname{definition}{Definition}{Definitions}
\crefname{equation}{Eq.}{Equations}
\crefname{table}{Table.}{Tables}
\crefname{figure}{Fig.}{Figures}
\crefname{section}{Section}{Sections}

\begin{document}


\twocolumn[

\aistatstitle{Still no free lunches: the price to pay for tighter PAC-Bayes bounds}

 \aistatsauthor{ Benjamin Guedj \And Louis Pujol }

 \aistatsaddress{ Inria and University College London \And  Université Paris-Saclay } ]

\begin{abstract}
  "No free lunch" results state the impossibility of obtaining meaningful bounds on the error of a learning algorithm without prior assumptions and modelling. Some models are expensive (strong assumptions, such as as subgaussian tails), others are cheap (simply finite variance). As it is well known, the more you pay, the more you get: in other words, the most expensive models yield the more interesting bounds. Recent advances in robust statistics have investigated procedures to obtain tight bounds while keeping the cost minimal. The present paper explores and exhibits what the limits are for obtaining tight PAC-Bayes bounds in a robust setting for cheap models, addressing the question: is PAC-Bayes good value for money?
\end{abstract}

\section{Introduction: about the "no free lunch" results}\label{sec:intro}

A class of results in statistics is known as ``no free lunch" statements \citep[][Chapter 7]{devroye2013probabilistic}. This kind of results deals with the fact that if one does not consider restrictions on the modelling of the data-generating process, one cannot obtain meaningful deviation bounds in a non-asymptotic regime. The well known tradeoff is that the more restrictive the assumptions, the tighter the bounds. Let us illustrate this classical phenomenon by a simple example.

Assume that we have a dataset consisting in $N$ real observations $x_1, \dots, x_N\in \mathbb{R}$ and consider they are independent, identically distributed (iid) realisations of a random variable following an unknown distribution $\mathrm{P}$. Our goal is to estimate the mean of $\mathrm{P}$ and build a confidence interval for this estimate. As a start, let us focus on the empirical mean, denoted by $\Bar{x}=\frac{1}{N}\sum_{i=1}^N x_i$. As ``no free lunch" results state, we have to consider a class of distributions to which $\mathrm{P}$ belongs.

A first type of restriction we can make can be called ``expensive models". Consider that $\mathrm{P}$ belongs to the class $\mathcal{P}_{\text{expensive}}^\sigma$ consisting of all probability distributions over $\mathbb{R}$ such that if $X$ follows distribution $\mathrm{P}$, then for all $\lambda \in \mathbb{R}$

$$ \mathbb{E}\left[ \exp\left\{\lambda(X-\mathbb{E}[X])\right\} \right] \leq \frac{\lambda^2 \sigma^2}{2}. $$

This class $\mathcal{P}_{\text{expensive}}^\sigma$ is known as the one of subgaussian random variables with variance factor $\sigma^2$ \citep[see][for a nice introduction to concentration theory]{boucheron2013concentration}. Let $\delta \in (0,1)$, 

\begin{equation}\label{eq:confidence1}
    \left[\,\, \Bar{x} \pm \frac{\sigma}{\sqrt{N}} \sqrt{2 \log\left( \frac{1}{\delta} \right)} \,\, \right]
\end{equation} 

is a confidence interval at level $1-\delta$ for the mean.

A second type of restriction can be called accordingly ``cheap models". Assume that the distribution $\mathrm{P}$ belongs to the class $\mathcal{P}_{\text{cheap}}^{\sigma}$, consisting of distributions with a finite variance, upper bounded by $\sigma^2$. Here Chebyshev's inequality straightforwadly gives us a confidence interval. Let $\delta \in (0,1)$, 

\begin{equation}\label{eq:confidence2}
    \left[\,\, \Bar{x} \pm \frac{\sigma}{\sqrt{N}} \sqrt{\frac{1}{\delta}} \,\, \right]
\end{equation} 

is a confidence interval at level $1-\delta$ for the mean. In that case, there is no hope to obtain significantly tighter confidence intervals if one uses the empirical mean \citep[as proved in][Proposition 6.2]{catoni2012challenging}.

\begin{figure}[t]
    \centering
    \includegraphics[width=.5\textwidth]{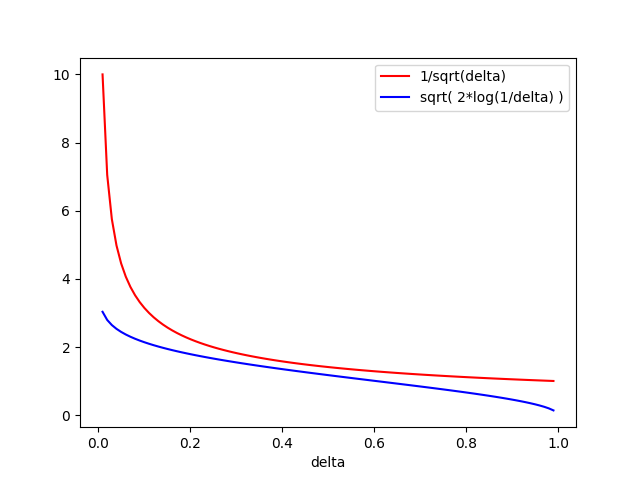}
    \caption{Relative sizes of confidence intervals: the dependence on $\delta$.}
    \label{fig1}
\end{figure}

Note that the dependence in $\delta$ is fairly different in both confidence intervals defined in \eqref{eq:confidence1} and \eqref{eq:confidence2}: for fixed $\sigma^2$ and $N$, the $\sqrt{2\log(1/\delta)}$ regime (the "good lunch") is obviously much more favorable than the $1/\sqrt{\delta}$ regime (the "bad lunch"). This is illustrated by \autoref{fig1}.

So while it is clear than the best confidence interval requires more stringent assumptions, there has been attemps at relaxing those assumptions -- or in other words, keep equally good lunches for a cheaper cost.

\paragraph{Organisation of the paper.} We provide an overview of recent advances in robust statistics (\cref{sec:robust}), and briefly introduce our notation (\Cref{sec:notation}) for PAC-Bayes learning (\Cref{sec:pacbayes}). We then propose in \Cref{sec:robustpacbayes} a detailed study on the structural limits which do not allow for PAC-Bayes bounds which are simultaneously tight and cheap. The paper closes with conclusive remarks in \Cref{sec:conclusion}.

\section{Robust statistics}\label{sec:robust}

Robust statistics adress the following question: is it possible to obtain a good lunch, with just a cheap model ? In the mean estimation case hinted in \Cref{sec:intro}, the question become: in the situation where $\mathrm{P} \in \mathcal{P}_{\text{cheap}}^{\sigma}$, can we build a confidence interval at level $1-\delta$ with a size proportional to $\frac{\sigma}{\sqrt{N}}\sqrt{2 \log(1/\delta)}$?

As mentioned above, there is no hope to achieve this goal with the empirical mean. Different alternative estimators have thus been considered in robust statistics, such as M-estimators \citep{catoni2012challenging} or median-of-means (MoM) estimators \citep[see][for a recent survey, and references therein]{lerasle2019lecture}.

The key idea of MoM estimators is to achieve a compromise between the unbiased but non-robust empirical mean and the biased but robust median. As before, let us consider a sample of $N$ real numbers $x_1, \dots ,x_N$, assumed to be an iid sequence drawn from a distribution $\mathrm{P}$. Let $K\leq N$ be a positive integer and assume for simplicity that $K$ is a divisor of $N$. To compute the MoM estimator, the first step consists in dividing the sample $(x_1,\dots,x_N)$ into $K$ distincts blocs $B_1, \dots , B_K$, each of length $N/K$. For each bloc we then compute the empirical mean

$$ \bar{x}_{B_i} = \frac{K}{N} \sum_{j \in B_i} x_j. $$

The MoM estimator is defined as the median of these means:
$$ \mathrm{MoM}_K(x_1 \dots, x_N) = \mathrm{median}\left\{ \bar{x}_{B_1}, \dots, \bar{x}_{B_K} \right\}. $$

This estimator have the following nice property.

\begin{proposition}[\cite{lerasle2019lecture}, Proposition 12]
Assume $\mathrm{P} \in \mathcal{P}_{\text{cheap}}^{\sigma} $, for $\delta = \exp\left(-\frac{K}{8} \right)$
\begin{equation}\label{eq:intervalmom}
    \left[ \mathrm{MOM}_K \pm \frac{\sigma}{\sqrt{N}} \times 4 \sqrt{2 \log\left(\frac{1}{\delta}\right)} \right]
\end{equation}
is a confidence interval for the mean at the level $1-\delta$.
\end{proposition}

This property is quite encouraging as for a cheap model we obtain a confidence interval similar, up to a numerical constant, to the best one \eqref{eq:confidence1} in \Cref{sec:intro}. However we also spot here an important limitation. The confidence interval \eqref{eq:intervalmom} for MoM is only valid for the particular error threshold $\delta = \exp\left(-K/8\right)$, which depends on the number of blocs $K$ (a parameter for the estimator $\mathrm{MoM}_K$). The estimator must be changed each time we want to evaluate a different confidence level.

An ever more limiting feature is that the error threshold $\delta$ is constrained and cannot be set arbitrarily small, as in \eqref{eq:confidence1} or \eqref{eq:confidence2}. Obviously, the number of blocks cannot exceed the sample size $N$, and the error threshold reaches its lowest tolerable value $\exp\left(-N/8\right)$. In other words, the interval defined in \eqref{eq:intervalmom} can have confidence at most $1-\exp\left(-N/8\right)$.

Is this strong limitation specific to MoM estimators? No, say \citet[][Theorem 3.2 and following remark]{devroye2016sub}. This limitation is universal: over the class $\mathcal{P}_{\text{cheap}}^{\sigma}$, there is no estimator $\hat{x}$ of the mean such that there exists a constant $L>1$ such that

$$ \left[ \hat{x} \pm \frac{\sigma}{\sqrt{N}} \times L \sqrt{2 \log\left(\frac{1}{\delta}\right)} \right] $$

is a confidence interval at level $1-\delta$ for $\delta$ lower than $e^{- \mathcal{O}(N)}$.

Overall, a good and cheap lunch is possible, at the extra price that the bound is no longer valid for all confidence levels.

\section{Notation}\label{sec:notation}

In the remainder of this paper, we focus on the supervised learning problem.  We collect a sequence of input-output pairs $(X_i, Y_i)_{i=1}^N \in (\mathcal{X}\times \mathcal{Y})^N$, which we assume to be $N$ independent realisations of a random variable drawn from distribution $\mathrm{P}$ on $\mathcal{X} \times \mathcal{Y}$. The overarching goal in statistics and machine learning is to select a hypothesis $f$ over a space $\mathcal{F}$ which, given a new input $x$ in $\mathcal{X}$, delivers an output $f(x)$ in $\mathcal{Y}$, hopefully close (in a certain sense) to the unknown true output $y$. The quality of $f$ is assessed through a loss function $\ell$ which characterises the discrepancy between the true output $y$ and its prediction $f(x)$, and we define a global notion of risk
$$ R(f) = \mathbb{E}_{(X,Y)\sim \mathrm{P}}\left[ \ell\left( f(X), Y \right) \right] .$$

As the expectation with respect to $\mathrm{P}$ is intractable, we need to resort to an estimator of the risk. The most intuitive and simple choice is the empirical risk, defined for each $f \in \mathcal{F}$ as

$$ R_N(f) = \frac{1}{N} \sum_{i=1}^N \ell( f(X_i), Y_i ).$$

In the following, we consider integrals over the hypotheses space $\mathcal{F}$. To keep notation as compact as possible we will write $\mu[g] = \int g\mathrm{d}\mu$ if $\mu$ is a measure over $\mathcal{F}$ and $g \in \mathcal{F}$ a $\mu$-integrable function.

\section{PAC-Bayes}\label{sec:pacbayes}

In this section, we briefly introduce the generalised Bayesian setting in machine learning, and the resulting generalisation bounds, the PAC-Bayesian bounds. PAC-Bayes is a sophisticated framework to derive new learning algorithms and obtain state-of-the-art generalisation bounds: as such, we are interested in studying how PAC-Bayes is compatible with good and cheap lunches. We refer the reader to \cite{guedj2019primer} for a recent survey on PAC-Bayes. We focus on bounds known in the PAC-Bayes literature based on the empirical risk as a risk estimator in two conditions corresponding to the ``expensive" and ``cheap"  models introduced in \Cref{sec:intro}.

\subsection{Generalised Bayes and PAC bounds}

The aim of machine learning is to find a good (in the sense of a low risk) hypothesis $f\in\mathcal{F}$. In the generalised Bayes setting, the learning algorithm does not output a single hypothesis but rather a \emph{distribution} $\rho$ over the hypotheses space $\mathcal{F}$.

The main advantage of PAC-Bayes over deterministic approaches which output single hypotheses (through optimisation of a particular criterion, model selection, etc.) is that distributions allow to capture uncertainty on hypotheses, and take into account correlations among possible hypotheses.

The quantity to control is then
$$ \rho[R] = \int_{\mathcal{F}} R(f) \mathrm{d}\rho(f) $$
which is an aggregated risk over the class $\mathcal{F}$ and represents the expected risk if the predictor $f$ is drawn from $\rho$ for each new prediction. The distribution $\rho$ is usually data-dependent and is referred to as a "posterior" distribution (by analogy with Bayesian statistics). We also fix a reference measure $\pi$ over $\mathcal{F}$, called the "prior" (for similar reasons). We refer to \cite{catoni2007pac} and \cite{guedj2019primer} for in-depth discussions on the choice of the prior.

% \subsection{PAC-Bayesian bouds}

The generalisation bounds associated to this setting are known as ``PAC-Bayesian" bounds, where PAC stands for Probably Approximately Correct. One important characteristic of PAC-Bayes bounds are that they hold true for any prior $\pi$ and posterior $\rho$. In practice, bounds are optimised with respect to $\rho$. In the following, we focus on establishing bounds for any choice of $\pi$ and $\rho$ and do not mean to optimise.

\subsection{Notion of divergence}

An important notion used in PAC-Bayesian theory is the divergence between two probability distributions \citep[see for example][for a survey on divergences]{csiszar2004information}. Let $\mathcal{E}$ be a measurable space and $\mu$ and $\nu$ two probability distributions on $\mathcal{E}$. Let $f$ be a nonnegative convex function defined on $\mathbb{R}+$ such that $f(1) = 0$, we define the $f$-divergence\footnote{We also use $f$ to denote hypotheses elsewhere in the paper, but we believe the context to always be clear enough to avoid ambiguity.} between $\mu$ and $\nu$ by

$$ \mathcal{D}_f(\mu, \nu) = \left\{
\begin{array}{ll}
\int f\left( \frac{\mathrm{d}\mu}{\mathrm{d}\nu} \right) \mathrm{d}\nu & \text{if $\mu \ll \nu$}, \\
+\infty & \text{otherwise.}
\end{array}
\right.
$$

Applying Jensen inequality we have that $\mathcal{D}_f(\mu, \nu)$ is always nonnegative and equal to zero if and only if $\mu = \nu$. The class of $f$-divergences includes many celebrated divergences, such as the Kullback-Leibler (KL) divergence, the reversed KL, the Hellinger distance, the total variation distance, $\chi^2$-divergences, $\alpha$-divergences, etc.

A divergence can be thought as a transport cost between two probability distributions. This interpretation will be useful for explaining PAC-Bayesian inequalities, where the divergence plays the role of a complexity term. In the following we will just use two types of divergence. The first is the Kullback-Leibler divergence and corresponds to the choice $f(x) = x \log x$, we denote it by

$$ \mathrm{KL}(\mu, \nu) = \left\{
\begin{array}{ll}
\int  \log\left( \frac{\mathrm{d}\mu}{\mathrm{d}\nu}\right) \mathrm{d}\mu & \text{if $\mu \ll \nu$}, \\
+\infty & \text{otherwise.}
\end{array}
\right.
$$

The second is linked to Pearson's $\chi^2$-divergence and corresponds to the choice $f(x) = x^2-1$. It is referred to as $\mathcal{D}_2$:

$$ \mathcal{D}_2(\mu, \nu) = \left\{
\begin{array}{ll}
\int \left( \frac{\mathrm{d}\mu}{\mathrm{d}\nu} \right)^2 \mathrm{d}\nu \  - 1 & \text{if $\mu \ll \nu$}, \\
+\infty & \text{otherwise.}
\end{array}
\right.
$$

To illustrate the behaviour of these two divergences, consider the case where $\mu$ and $\nu$ are normal distribution on $\mathbb{R}^d$.

\begin{proposition}
If $\mathcal{E} = \mathbb{R}^d$, $\mu = \mathcal{N}(a, I)$ and $\nu = \mathcal{N}(0, I)$ (where $I$ stands for the $d\times d$ identity matrix), we have

$$ \left\{
\begin{array}{l}
\mathcal{D}_2(\mu, \nu) = e^{\Vert a \Vert^2} -1, \\
 \mathrm{KL}( \mu, \nu) = \frac{1}{2}\Vert a \Vert^2.
 \end{array}
\right.
$$

\end{proposition}

We therefore see that the divergence $\mathcal{D}_2$ penalises much more the gap between both distributions than the Kullback-Leibler divergence.

\subsection{Expensive PAC-Bayesian bound}

The first PAC-Bayesian bound we present is called ``expensive PAC-Bayesian bound" in the spirit of \Cref{sec:intro}: it is obtained under a subgaussian tails assumption. More precisely, we suppose here that for every $f \in \mathcal{F}$, the distribution of the random variable $\ell( f(X), Y )$ belongs to $\mathcal{P}_{\text{expensive}}^\sigma$, which means

$$ \mathbb{E}\left[ \exp\left\{\lambda( \ell(f(X), Y) - R(f) )\right\} \right] \leq \frac{\lambda^2 \sigma^2}{2}, \quad \forall \lambda \in \mathbb{R}. $$

In this situation we have the following bound, close to the ones obtained by \cite{catoni2007pac}. 

\begin{proposition}

Assume that for any $f \in \mathcal{F}$, $\ell( f(X), Y ) \in \mathcal{P}_{\text{expensive}}^\sigma$. For any prior $\pi$, posterior $\rho$ and any $\delta \in (0,1)$, the following inequality holds true with probability greater than $1-\delta$
$$ \rho[R] \leq \rho[R_N] + \frac{\sigma}{\sqrt{N}} \sqrt{ 2 \left( \log \left(\frac{1}{\delta}\right) + \mathrm{KL}(\rho, \pi) \right) } .$$

\end{proposition}

\begin{proof}

The proof can be decomposed into two steps. The first is to use the following lemma, consisting in a change of measure between the posterior and the prior.

\begin{lemma}[\cite{csiszar1975divergence} -- \cite{boucheron2013concentration}, Corollary 4.15]

Let $g$ be a measurable function $g : \mathcal{F} \mapsto \mathbb{R}$ such that $\pi\left[ e^{g} \right]$ is finite. The following inequality holds true
$$ \rho[g] \leq \log \pi \left[ e^g \right] + \mathrm{KL}(\rho, \pi). $$

\end{lemma}

Let $\lambda$ be a positive number and applying this result for the function $\lambda(R-R_N)$:
$$ \rho[R] \leq \rho[R_N] + \frac{1}{\lambda} \left( \log \pi \left[ e^{\lambda(R-R_N)} \right] + \mathrm{KL}(\rho, \pi)  \right). $$

The next step is to control $\log \pi \left[ e^{\lambda(R-R_N)} \right]$ in high probability. With probability $1-\delta$ we have, by Markov's inequality
$$ \pi\left[ e^{ \lambda(R-R_N) } \right] \leq \frac{\mathbb{E}\left[ \pi\left[ e^{ \lambda(R-R_N) } \right]  \right]}{\delta}. $$

By Fubini's theorem we can exchange the symbols $\mathbb{E}$ and $\pi$. Using the assumption $\mathcal{P}_{\text{expensive}}^\sigma$, we obtain with probability greater than $1-\delta$

$$ \pi\left[ e^{ \lambda(R-R_N) } \right] \leq \frac{ \exp\left\{ \lambda^2 \sigma^2 / 2N  \right\}}{\delta}. $$

Now, putting these results together and setting $$\lambda = \frac{\sqrt{2N \left( \log\left( \frac{1}{\delta} \right) + \mathrm{KL}(\rho, \pi) \right)}}{\sigma}$$ we obtain the desired bound.
\end{proof}

A PAC-Bayesian inequality is a bound which treats the complexity in the following manner:
\begin{itemize}
    \item At first, a global complexity measure is introduced with the change of measure and is characterised by the divergence term, measuring the price to switch from $\pi$ (the reference distribution) to $\rho$ (the posterior distribution on which all inference and prediction is based);
    \item Next, the stochastic assumption on the data-generating distribution is used to control $\pi\left[ e^{ \lambda(R-R_N) } \right]$ with high probability.
\end{itemize}

\subsection{Cheap PAC-bayesian bound}

The vast majority of works in the PAC-Bayesian literature focuses on expensive model. The main reason is that it include the situation where the loss $\ell$ is bounded, a common assumption in machine learning. The case where $\ell( f(X, Y )$ belongs to a cheap model has attracted far less attention: recently, \cite{alquier2018simpler} have obtained the following bound.

\begin{proposition}[\cite{alquier2018simpler}, Theorem 1]

Assume that for any $f \in \mathcal{F}$, $\ell( f(X), Y ) \in \mathcal{P}_{\text{cheap}}^{\sigma}$. For any prior $\pi$, posterior $\rho$ and any $\delta \in (0,1)$, the following inequality holds is true with probability greater than $1-\delta$
$$ \rho[R] \leq \rho[R_N] + \frac{\sigma}{\sqrt{N}} \sqrt{ \frac{\mathcal{D}_2(\rho, \pi)+1}{\delta} }. $$

\end{proposition}

The proof \citep[see][]{alquier2018simpler} uses the same elementary ingredients as in the expensive case, replacing the Kullback-Leibler divergence by $\mathcal{D}_2$ and the dependence in $\delta$ moves from $\sqrt{ 2 \log(1/\delta) }$ to $\frac{1}{\sqrt{\delta}}$. Note the correspondence between these two bounds and the confidence intervals introduced in \Cref{sec:intro}.

\section{A good cheap lunch: towards a robust PAC-Bayesian bound?}\label{sec:robustpacbayes}

If we take a closer look at the aforementioned PAC-Bayesian bounds with a robust statistics viewpoint, the following question arises: \textbf{can we obtain a PAC-Bayesian bound with a $\sqrt{\log (1/\delta) }$ dependence (possibly up to a numerical constant) in the confidence level with the cheap model?} In this section we shed light on some structural issues. In the following, we assume the existence of $\sigma > 0$ such that for every $f \in \mathcal{F}$, $\ell( f(X), Y ) \in \mathcal{P}_{\text{cheap}}^{\sigma}$.

\subsection{A necessary condition}

Let $\widehat{R}$ be an estimator of the risk. Here is a prototype of the inequality we are looking for: for any $\delta \in (0,1)$, with probability $1-\delta$

$$ \rho[R] \leq \rho\left[\widehat{R}\right]+ \frac{\sigma}{\sqrt{N}} \mathrm{A}( \rho, \pi, \delta ),$$

where

$$ \mathrm{A}( \rho, \pi, \delta ) \underset{\delta \rightarrow 0}{=} \mathcal{O}\left(\sqrt{\log(1/\delta)} \right) . $$

If we choose $\rho = \pi = \delta_{\{f\}}$ (Dirac mass in the single hypothesis $f$), the existence of such a PAC-Bayesian bound valid for all $\delta$ implies that
$$\left[ \widehat{R}(f) \pm \frac{\sigma}{\sqrt{N}} \times c \sqrt{\log (1/ \delta)} \right] $$

is a confidence interval for the risk $R(f)$ for any level $1 - \delta$, where $c$ is a constant.

Thus, a necessary condition for a PAC-Bayesian bound to be valid for all risk level $\delta$, is to have tight confidence intervals for each $f \in \mathcal{F}$.

However, as covered in \Cref{sec:robust}, such estimators do not exist over the class $\mathcal{P}_{\text{cheap}}^\sigma$, and the possibility to derive tight confidence interval is limited by the fact that the level $\delta$ must be greater that a positive constant of the form $e^{-\mathcal{O}(N)}$.

\subsection{A $\delta$-dependant PAC-bayesian bound?}

As a consequence, there is simply no hope for a robust PAC-Bayesian bound valid for any error threshold $\delta$, for essentially the same reason which prevents it in the mean estimation case. The question we address now is the possibility of obtaining a robust PAC-Bayesian bound, with a dependence of magnitude $\sqrt{ 2 \log(1/\delta) }$ (possibly up to a constant), with a possible limitation on the error threshold $\delta$. In the following we assume to have a risk estimator $\widehat{R}$ and an error threshold $\delta > 0$ such that there exists a constant $C>0$ such that for all $f \in \mathcal{F}$,

$$ \left[ \widehat{R}(f) \pm \frac{\sigma}{\sqrt{N}} \times C\ \sqrt{\log(1/\delta)} \right] $$

is a confidence interval for $R(f)$ at level $1-\delta$. MoM is an example of such estimator. Let us stress that $\delta$ is fixed and cannot be used as a free parameter.

As seen above, a PAC-Bayesian bound proof proceeds in two steps:
\begin{itemize}
    \item First, we use a convexity argument to control the target quantity $\rho[  R- \widehat{R} ]$ by an upper-bound involving a divergence term and a term of the form $ g^{-1} \left( \pi\left[ g(R-\widehat{R}) \right] \right) $ where $g$ is a nonnegative, increasing and convex function;
    \item Second, we control the term $\pi\left[ g(R-\widehat{R}) \right]$ in high probability, using Markov's inequality.
\end{itemize}

The first step does not require any use of a stochastic model on the data, and is always valid, regardless of whether we have a cheap or an expensive model. The second step uses the model and introduce the dependence in the error rate $\delta$ on the right-term of the bound: $g^{-1}( 1 / \delta )$. In the case of the ``expensive bound", we had $g = \exp$, and the dependence was $ \log( 1/\delta ) $, the final rate  $\sqrt{ \log( 1/\delta ) }$ was obtained by choosing a relevant value for $\lambda$.

Let us follow this scheme to obtain a robust PAC-Bayesian bound. The first step gives

$$ \rho[R] \leq \rho[\widehat{R}] + \frac{1}{\lambda} \left( \log \pi \left[ e^{\lambda(R-\widehat{R})} \right] + \mathrm{KL}(\rho, \pi)  \right). $$

Our goal is now to control $\pi \left[ e^{\lambda(R-\widehat{R})} \right]$ in high probability. Let us see why it seems impossible.

\subsubsection{The case $\pi = \delta_{\{f\}}$}

Let us start with a very special case, where the prior is a Dirac mass on some hypothesis $f \in \mathcal{F}$. Then  $$\frac{1}{\lambda}\log \pi \left[ e^{\lambda(R-\widehat{R})} \right] = R(f)-\widehat{R}(f).$$ Using how $\widehat{R}$ is defined we can bound this quantity in the following way: with probability $1 - \delta$,
$$ R(f)-\widehat{R}(f) \leq \frac{\sigma}{\sqrt{N}} \times C \sqrt{ \log(1/ \delta) }. $$

Another way to formulate this result is to say that there exists an event $\mathcal{A}_f$ with probability greater than $1-\delta$ such that for all $\omega \in \mathcal{A}_f$, the following holds true:
$$ (R(f)-\widehat{R}(f, \omega)) \leq \frac{\sigma}{\sqrt{N}} \times C \sqrt{2 \log(1/ \delta) }. $$

In this example, we can control $\log \pi \left[ e^{\lambda(R-\widehat{R})} \right]$, at the price of a maximal constraint on the choice of the posterior. Indeed, the only possible choice for $\rho$ for the Kullback Leibler $\mathrm{KL}(\rho, \pi)$ to make sense is $\rho=\pi=\delta_{\{f\}}$.

\subsubsection{The case $\pi = \alpha \delta_{\{f_1\}} + (1-\alpha) \delta_{\{f_2\}}$}

Consider now a somewhat more sophisticated choice of prior which is a mixture of two Dirac masses in two distinct hypotheses. We do not fix the mixing proportion $\alpha$ and allow it to move freely between $0$ and $1$. The goal is to control the quantity
$$ \pi \left[ e^{\lambda(R-\widehat{R})} \right] = \alpha e^{\lambda(R(f_1)-\widehat{R}(f_1))} + (1-\alpha)e^{\lambda(R(f_2)-\widehat{R}(f_2))} .$$ More precisely, for all $\alpha \in (0,1)$, we want to find an event $\mathcal{A}_{\alpha}$ on which this quantity is under control. In view of the prior's structure, the only way to ensure such a control is to have $\mathcal{A}_{\alpha} \subset \mathcal{A}_{f_2}\cap \mathcal{A}_{f_2}$, where $\mathcal{A}_{f_1}$ (resp. $\mathcal{A}_{f_2}$) is the favourable event for the concentration of $\widehat{f_1}$ (resp. $\widehat{f_2}$) around its mean. 

By the union bound, we have that with probability greater than $1 - 2 \delta$

$$\frac{1}{\lambda} \log \pi \left[ e^{\lambda(R-\widehat{R})} \right] \leq \frac{\sigma}{\sqrt{N}} \times C \sqrt{ \log(1/ \delta) }.$$

We have a double problem here. As above, if we want the final bound to be non-vacuous, we have to ensure that $\mathrm{KL}(\rho, \pi)$ is finite, which restricts the support for the posterior to be included in the set $\{ f_1, f_2 \}$. In addition, the probability with which we can guarantee the PAC-Bayesian bound is now $1-2\delta$... 

\subsubsection{Limitation}

... which hints at the fact that this will become $1-K\delta$ if the support for the prior contains $K$ distinct hypotheses. If $K \geq 1/\delta$, the bound becomes vacuous. In particular, we cannot obtain a relevant bound using this approach in the situation where the cardinal of $\mathcal{F}$ is infinite (which is commonly the case in most PAC-Bayes works).

This limiting fact highlights that to derive PAC-Bayesian bounds, we cannot rely on the construction of confidence interval for all $R(f)$ for a fixed error threshold $\delta$. The issue is that when we want to transfer this local property into a global one (valid for any mixture of hypotheses by the prior $\pi$), we cannot avoid a worst-case reasoning by the use of the union bound.

The established bound in PAC-Bayesian literature, both in cheap and expensive models, repeatedly use the fact that when we assume that for any $f \in \mathcal{F}$,

$$ \log \mathbb{E}\left[ e^{\lambda(R(f)-\ell(f(X), Y))} \right] \leq \frac{\lambda^2\sigma^2}{2}, \  \forall \lambda \in \mathbb{R} $$

or

$$ \mathrm{var}\left( \ell( f(X), Y ) \right) \leq \sigma^2,$$

we make an implicit assumption on the integrability  of the tail of the distribution of $\ell(f(X), Y)$. This argument is crucial for the second step of the PAC-Bayesian proof because, by Fubini's theorem, it allows to convert a local property (the tail distribution of each $\ell( f(X), Y )$) into a global one (the control of $\pi\left[ e^{\lambda(R-R_N)} \right]$ or $\pi\left[ (R-R_N))^2 \right]$ in high probability). 

\subsection{Is there yet another path for hope?}

We have identified a structural limitation to derive a tight PAC-bayesian bound in a cheap model. We make the case that we cannot replicate the PAC-Bayesian proof presented in \Cref{sec:pacbayes}. To conclude this section, we want to highlight the fact that, up to our knowledge, no proof of PAC-Bayesian bounds avoid these two steps \citep[see for example the general presentation in][]{begin2016pac}.

What if we try to avoid the change of measure step and try to control directly $\rho[R] - \rho[\widehat{R}]$ in high probability ? We remark that $\rho$ can only be chosen with the information given by the observation of $\widehat{R}(f)$, where $f\in\mathcal{F}$. In particular we cannot obtain any information of the concentration of each $\widehat{R}(f)$ around $R(f)$ as such a knowledge requires to know the true risk. So it seems that a direct control cannot avoid starting as a "worst-case" bound:
$$ \rho[R] - \rho[\widehat{R}] \leq \sup_{f \in \mathcal{F}} \left\{ R(f) - \widehat{R}(f)  \right\} . $$

Then we have to control $\sup_{f \in \mathcal{F}} \left\{ R(f) - \widehat{R}(f)  \right\}$ in high probability (see \citealp{van1996weak} for a general presentation on such controls, and \citealp{lerasle2019lecture} for recent results in the special case where $\widehat{R}$ is a MoM estimator). However the obtained bound will take the following prototypic form:
$$ \rho[R] \leq \rho[\widehat{R}] + \text{complexity term},$$

where the complexity term does not depend on the distribution $\rho$. Thus the optimisation of the right term leads to choose $\rho$ as the Dirac mass in $\underset{f \in \mathcal{F}}{\arg \min}\, \widehat{R}(f)$.

So the overall procedure amounts to a slightly modified empirical risk minimisation (where the empirical mean is replaced with any estimator of the risk), and will not fall into the category of generalised Bayesian approaches which take into account the uncertainty on hypotheses. We would therefore loose pretty much all the strengths of PAC-Bayes. 

\section{Conclusion}\label{sec:conclusion}

The present paper contributes a better understanding of profound structural reasons why good cheap lunches (tight bounds under minimal assumptions) are not possible with PAC-Bayes, by walking gently through elementary examples.

From a theoretical perspective, PAC-Bayesian bounds requires too strong assumptions to adapt robust statistics results (where almost good lunches can be obtained for cheap models -- with the limitation that the confidence level is constrained). The second step of the proof we have shown requires to transform a local hypothesis, a control of some moments of $\ell( f(X), Y )$ into a global one, valid for all mixture of hypotheses by the prior $\pi$. As covered above,  this transformation seems impossible.

To close on a more positive note after this negative result, let us stress that even if it does not seem possible to conciliate PAC-Bayes and robust statistics, we believe that recent ideas from robust statistics could be used in practical algorithms inspired by PAC-Bayes. In particular, we leave as an avenue for future work the empirical study of PAC-Bayesian posteriors (such as the Gibbs measure defined as $\rho \propto \exp(-\gamma\widehat{R})\pi$ for any inverse temperature $\gamma>0$) where the risk estimator is not the empirical mean (as in most PAC-Bayes works) but rather a robust estimator, such as MoM.

% \subsubsection*{Acknowledgements}

\bibliographystyle{plainnat}
\bibliography{biblio}

\end{document}